\documentclass{svproc}

\usepackage{url}

\usepackage[utf8]{inputenc}
\usepackage[T1]{fontenc}

\usepackage{amsmath,amsfonts,amssymb,mathtools}
\usepackage{microtype}
\usepackage[subtle]{savetrees}
\usepackage{xcolor}
\usepackage[normalem]{ulem} 
\usepackage{enumitem}
\usepackage{multicol}

\usepackage{graphicx}
\graphicspath{ {./figures/} }

\renewcommand{\baselinestretch}{0.90}
\input{abkuerzung.sty}

\begin{document}
	\mainmatter              	
	\title{A Concise Overview of Safety Aspects \\ in Human-Robot Interaction\thanks{This work was supported by the Bavarian State Ministry for Economic Affairs, Regional 
	Development and Energy (StMWi) as part of the project SafeRoBAY (grant number: DIK0203/01), the European Union’s Horizon 2020 research and innovation programme as part of the 
	project DARKO (grant no. 101017274), and the Lighthouse Initiative Geriatronics by StMWi Bayern (Project X, grant no. IUK-1807-0007// IUK582/001) and LongLeif GaPa gGmbH 
	(Project Y). \\[1cm] 
    {\small Accepted for Human-Friendly Robotics 2023: 16th International Workshop.} 
    }}
	\titlerunning{Safety Aspects in Human-Robot Interaction}  %
	\author{Mazin Hamad \and Simone Nertinger \and Robin J. Kirschner \and \\ Luis Figueredo \and Abdeldjallil Naceri \and Sami Haddadin}
	\authorrunning{Mazin Hamad et al.} %
	\tocauthor{Mazin Hamad, Simone Nertinger, Robin J. Kirschner, Luis Figueredo, Abdeldjallil Naceri, and Sami Haddadin}
	\institute{
		Chair of Robotics and Systems Intelligence,\\ 
		Munich Institute of Robotics and Machine Intelligence, \\
		Technical University of Munich, DE-80992 Munich, Germany,\\
		\email{mazin.hamad@tum.de}, %
		\texttt{https://www.ce.cit.tum.de/rsi/team/hamad-mazin/}
	}
	
	\maketitle              %
	
	\begin{abstract}
		As of today, robots exhibit impressive agility but also pose potential hazards to humans using/collaborating with them. Consequently, safety is considered the most paramount factor in human-robot interaction (HRI). This paper presents a multi-layered safety architecture, integrating both physical and cognitive aspects for effective HRI. We outline critical requirements for physical safety layers as service modules that can be arbitrarily queried. Further, we showcase an HRI scheme that addresses human factors and perceived safety as high-level constraints on a validated impact safety paradigm. The aim is to enable safety certification of human-friendly robots across various HRI scenarios.
	\end{abstract}
	
	\section{Introduction}
	\label{sec:intro}
	
	Human-friendly robots are distinguished by their ability to delicately react and physically interact with the world through compliant hardware and adaptive controllers \cite{haddadin2016physical}. However, despite significant advances in their tactile design, robots in the real world are still hardly deployed for close collaborative tasks together with humans. 
	Among the many challenges facing real-world human-robot interaction (HRI), physical safety is often considered the most pressing one. Moreover, in order to be accepted and deployed in close and effective interaction with human users, an intelligent robotic assistant must surpass the mere criteria of being contact-free and stress-free, i.e., physically safe. A human-friendly robot is required to be gracefully safe (GS), which we define as both possessing and exhibiting a (i) feasible, (ii) time-efficient, (iii) comfortable, and (iv) intuitive behaviour (\ie, perceived to be natural by the human user/coworker), while simultaneously being always human-safe. The concept of \emph{graceful robot behaviour} was originally introduced in \cite{gulati2008high} as being safe, fast, comfortable, and intuitive.
	However, such gracefulness should be further emphasized by ensuring safe robot behaviour in shared and collaborative spaces with humans, ultimately allowing for safety certification. This means the movements of the involved assistive robots should be physically as well as psychologically safe while additionally considering the efficacy of the human-robot team. Robots with such features are hereby termed \emph{gracefully safe robots}.  \looseness=-1
	
	Graceful robot navigation and reactive motion control strategies have been gaining momentum recently, where they have been shown to directly influence the quality and efficiency of HRI \cite{park2016graceful,moreno2020kinodynamic,haviland2022holistic,haviland2021neo}. Nonetheless, to enable physical human-robot interaction (pHRI) in real-world scenarios \cite{wang2020overview}, safety standards are decisive \cite{valori2021validating}. They govern the mechanical design, motion planning, and low-level control aspects of human-friendly robots in both industrial and domestic/service spaces. 
	To adhere to these standards, 
	a semi-automated, temporal logic-based risk analysis methodology for collaborative robotic applications that relies on formal verification techniques was introduced in \cite{vicentini2019safety}. 
	Furthermore, fundamental research about collisions and their consequences has received considerable attention from the robotics community. Concerning the safety of physical contacts, unintended robot-to-human impact scenarios are classified into five main contact scenarios~\cite{haddadin2009dlr_PartI}. Besides clamping in the robot structure, these include free, constrained, partially constrained, and secondary impacts. For scenarios involving desired contacts, such as hand-over tasks, smooth minimal-jerk movements on the robot side are known to improve the overall performance of the collaborative task with the human partner \cite{huber2008human}. Moreover, jerky/oscillatory motions are typically uncomfortable or even hazardous for people with specific conditions such as spinal cord injuries \cite{gulati2008high}. In addition to physical integrity,  the robot's behaviour plays a critical role in psychological safety. For instance, unexpected robot motion behaviours have been shown to trigger involuntary motions of users as a reaction of startle and surprise \cite{kirschner2022expectable}. In a similar fashion, any changes to the underlying functional modes of the collaborative robot, and consequently its applied motion commands, should be smooth to ensure that the interaction is executed efficiently and pleasantly \cite{svarny2022functional}. 
	
	Even though many building blocks and features for safe HRI exist \cite{park2016graceful,moreno2020kinodynamic,haviland2022holistic,haviland2021neo}, all these solutions still need to be integrated with recent concepts of graceful robot motion. 
	However, little attention is being paid to safety architectures that enable adequate simultaneous treatment of gracefulness and human-friendliness requirements of HRI scenarios. 
	This work aims to fill this fundamental gap hampering real-world HRI deployment. Herein, we propose a framework for gracefully safe robots, which in addition to physical and psychological safety connected to graceful features, addresses additional implementation hurdles \cite{tapus2007grand,eder2014towards,andreasson2015autonomous,madsen2015integration,koopman2017autonomous,kortner2016ethical,korchut2017challenges,dudek2019cyber,holland2021service,salvini2022safety}
	and allows for further integration of other critical challenges (such as, e.g.,  scalable integration, efficient coordination, dynamic mobile manipulation, optimal environment perception/sensing, purposeful communication, risk assessment, and decision making), as well as societal and ethical concerns (including data privacy and personal security \cite{torresen2018review,akalin2019evaluating,saplacan2021ethical}).
	
	\section{Problem Statement and Contribution}
	\label{sec:problem_contribution}
	As of today, a couple of solutions exist for different physical and cognitive safety aspects of HRI \cite{hamad2023modularize,kim2021human,figueredo2021planning,chen2018planning,kirschner2022expectable,nertinger2022acceptance,tapus2011user}. However, fulfilling the strict safety requirements of collaborative robotic systems while maintaining adequate graceful and human-friendly behaviour is still a significant challenge that has yet to be fully overcome. 
	To tackle this, we define the \textit{gracefully safe (GS)} behaviour for human-friendly robots by adopting and reinterpreting the original definition of being graceful in \cite{gulati2008high} as follows. Firstly, we clarified the safety requirement of the graceful robot behaviour as being related to motion constraints. In other words, by \emph{safe} in \cite{gulati2008high} it was rather meant that the robot motion fulfills the governing constraints (i.e., feasible). Secondly, we modified the gracefulness characteristic of being \emph{fast} to \emph{time-efficient} since the involvement of human safety aspects may pose different objectives on the human-robot collaborative task execution. Thirdly, as two characteristics of a graceful robot behaviour (namely, being comfortable and intuitive to human users) are inherent to perceived safety and acceptance, an independent comprehensive safety framework can be employed to tackle those requirements. As a quid pro quo, the task execution pipeline of the robot, which includes the motion controllers, motion planners, and task planners, must be reactive and adaptable, (i.e.,
	capable of addressing time constraints and additional costs imposed by human safety requirements).   
	
	Frameworks to achieve a GS behaviour, and further enable safety certification of HRI applications, should be suitably designed to simultaneously integrate the most prominent results concerning various physical and cognitive safety aspects from one side with robot motion planning and low-level control on the other. In addition, this synergy must be achieved as prescribed at the task planning and interaction dynamics level, where safe performance trade-offs between being very \textit{conservative} towards safety or \textit{just-as-needed} to improve the productivity of the interactive task can also be incorporated. 
	In this paper, we systematically tackle the central missing link to overcome the aforementioned gaps by proposing a \emph{multi-layered architecture for addressing safety aspects of human-friendly robots during HRI scenarios in both industrial and domestic settings}. 
	Overall, the main contributions of this article can be summarized as follows.
	\begin{itemize}[topsep=2pt]
		\item Based on an extensive literature analysis of multiple HRI dimensions, we distinguish between various physical and cognitive safety aspects that must be simultaneously fulfilled by human-friendly robots during GS-HRI;
		\item We identify instantaneous inputs/outputs and resource requirements of each physical safety layer;    
		\item Further, we detail the impact safety layer, showing how it can be implemented at the robot task planning and motion/control level. For this, we propose the so-called \emph{Safety-as-a-Service} service concept as an integrated multi-layered architecture for comprehensive safety consideration in HRI; 
		\item Finally, with the help of some initial integration results, we discuss how cognitive safety layers can be implemented on top of the physical ones in the design of GS-HRI. \looseness=-1
	\end{itemize}

	\section{Proposed Multi-layered HRI Safety Architecture}
	\label{sec:safe_HRI_architecture}
	For HRI applications, safety and security are among the most critical dimensions to consider. The term 'safety' typically refers to potential physical harm, whereas the term 'security' broadly refers to many aspects related to health, well-being, and aging \cite{akalin2019evaluating}. Consequently, investigating safety aspects for graceful HRI requires a multidisciplinary perspective. Typically, HRI safety aspects can be divided into physical and perceived safety, with the latter being an under-addressed topic in the robotics literature \cite{akalin2022you}. %
	
	We carried out a focused literature review to identify the following critical physical and cognitive safety aspects, which must be simultaneously considered by human-friendly robots for a GS-HRI 
	
	\setlength{\multicolsep}{6.0pt plus 2.0pt minus 1.5pt}%
	\setlength\columnsep{2.5pt}
	\begin{multicols}{3}
		\begin{itemize}
			\item Impact safety
			\item Acceptance
			\item Ergonomics
			\item Perceived safety
			\item Musculoskeletal safety
			\item Personalization
		\end{itemize}
	\end{multicols}
	
\begin{figure}%
		\centering
		\includegraphics[width=0.7\textwidth]{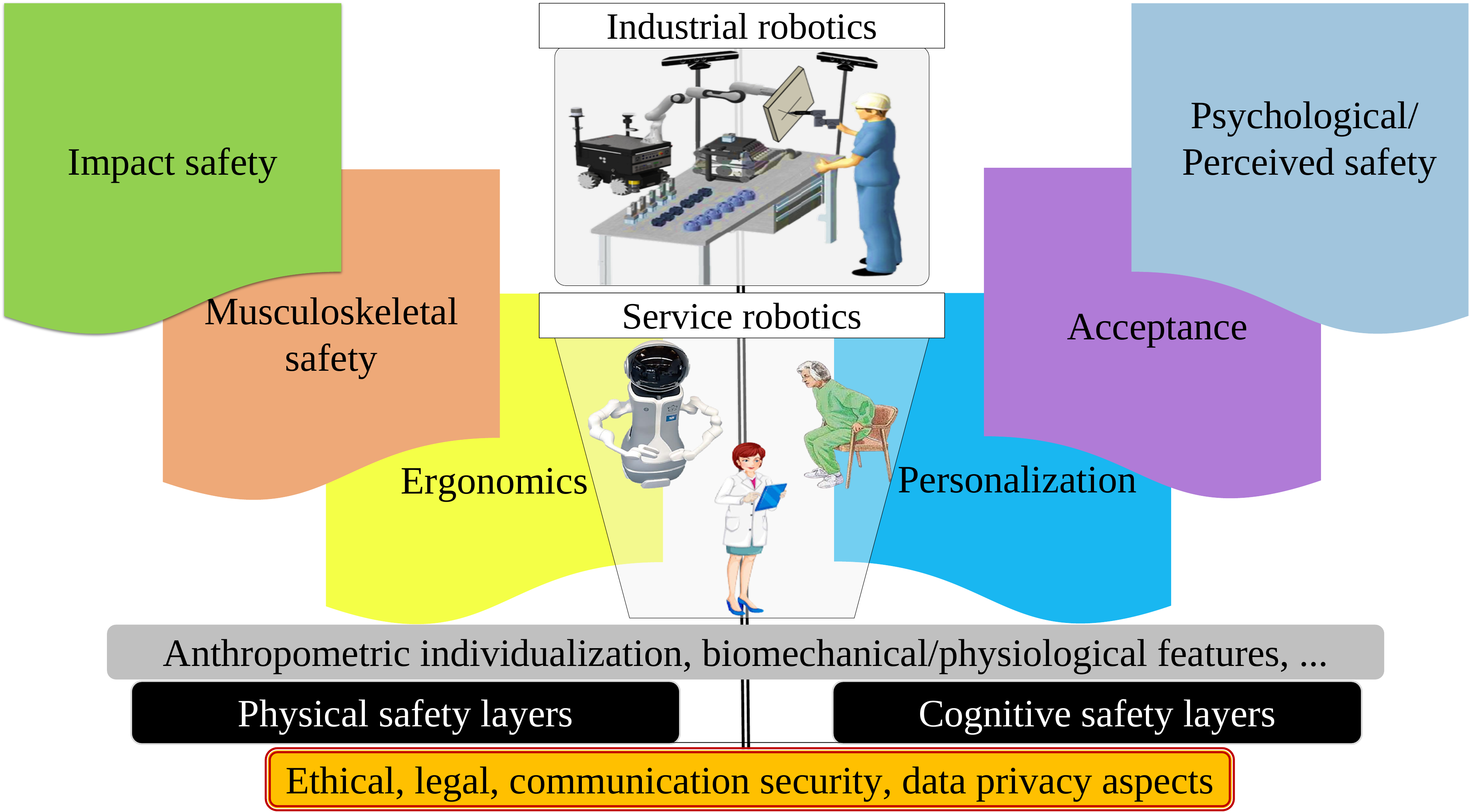}
		\caption[HRI safety layers]{HRI safety layers in industrial and service settings. On top of ethical, legal, and security aspects, we distinguish between physical and cognitive safety layers; both are subject to anthropomorphic personalization and various user-related customizations.}
		\label{fig:safety_layers}
	\end{figure}

	Based on that, we propose a multi-layered architecture for addressing safety aspects in HRI scenarios in both industrial and service/domestic settings, see Fig.~\ref{fig:safety_layers}. 
	
	\subsection{Identified safety layers for GS-HRI}
	\label{sec:safety_layers}
	
	Following an in-depth, focused literature review process, we identified the following key physical and cognitive safety layers that altogether cover the main aspects to be simultaneously considered for a gracefully safe and human-friendly robotic behaviour during HRI.  
	
	\subsubsection{Impact safety}
	\label{sec:impact_safety}
	Since contact is unavoidable and even desired in many applications, several studies, mostly employing cadavers and other human surrogates in addition to volunteers, have focused on understanding the pain thresholds and injury mechanisms of several human body parts to delimit the injurious conditions 
	\cite{yamada1997evaluation,muttray2014collaborative,haddadin2009requirements,haddadin2012understand,povse2016tool,behrens2014study,fraunhofer2019human}. 
	Important to notice is that most of the impact experiments reported in the literature were typically conducted on human cadavers from older adult subjects. For instance, Kent et al. \cite{kent2005structural} pointed out that overly large confidence intervals are produced on injury risk assessments in impact studies done with cadavers from older adults (as compared to those of young adults). Consequently, several researchers tried to overcome this problem by investigating the effect of age on the injury tolerance of humans and hence, developing some scaling laws \cite{crandall2011human}. Moreover, previous research has indicated that, on average, males experience less bone loss and slower cortical thinning rate than females as they age \cite{seeman2001during,lillie2016evaluation}. 
	Several biomechanical limits were proposed for the safety of robotic impact against humans, and the insights from biomechanical injury analysis were already imported into robotics \cite{haddadin2015physical}. Furthermore, the theoretical concepts behind the proposed pain/injury biomechanics-based paradigm have influenced many safety requirements stated in standardization documents such as EN ISO 13482 for personal care robots~\cite{robots2014robotic}, EN ISO 10218-1 and -2, as well as ISO/TS 15066 for industrial collaborative robots~\cite{ISO10218,TS15066}.  
	In addition to mitigating the involved human injury risks at the post-contact phase of the collision event pipeline \cite{haddadin2017survey}, pre-collision strategies are also required for a safe operation around humans in shared workspaces \cite{kulic2007pre}. 
	A comprehensive dummy crash test-based assessment of human injury risks when colliding with personal mobility devices and service robots was recently conducted in \cite{paez2022crash}. Comparing the risks faced by different pedestrian categories, it was shown that multiple serious injuries due to collisions could occur when the speeds exceed a certain threshold. Additionally, severe head injuries from falling to the ground after the initial impact were predicted from the secondary impact analysis. To reduce the impact injury risks in both cases, the authors suggested using absorbent materials or lowering the differential speed at impact as mitigation strategies. 
	
\begin{figure}
		\centering
		\includegraphics[width=0.95\textwidth]{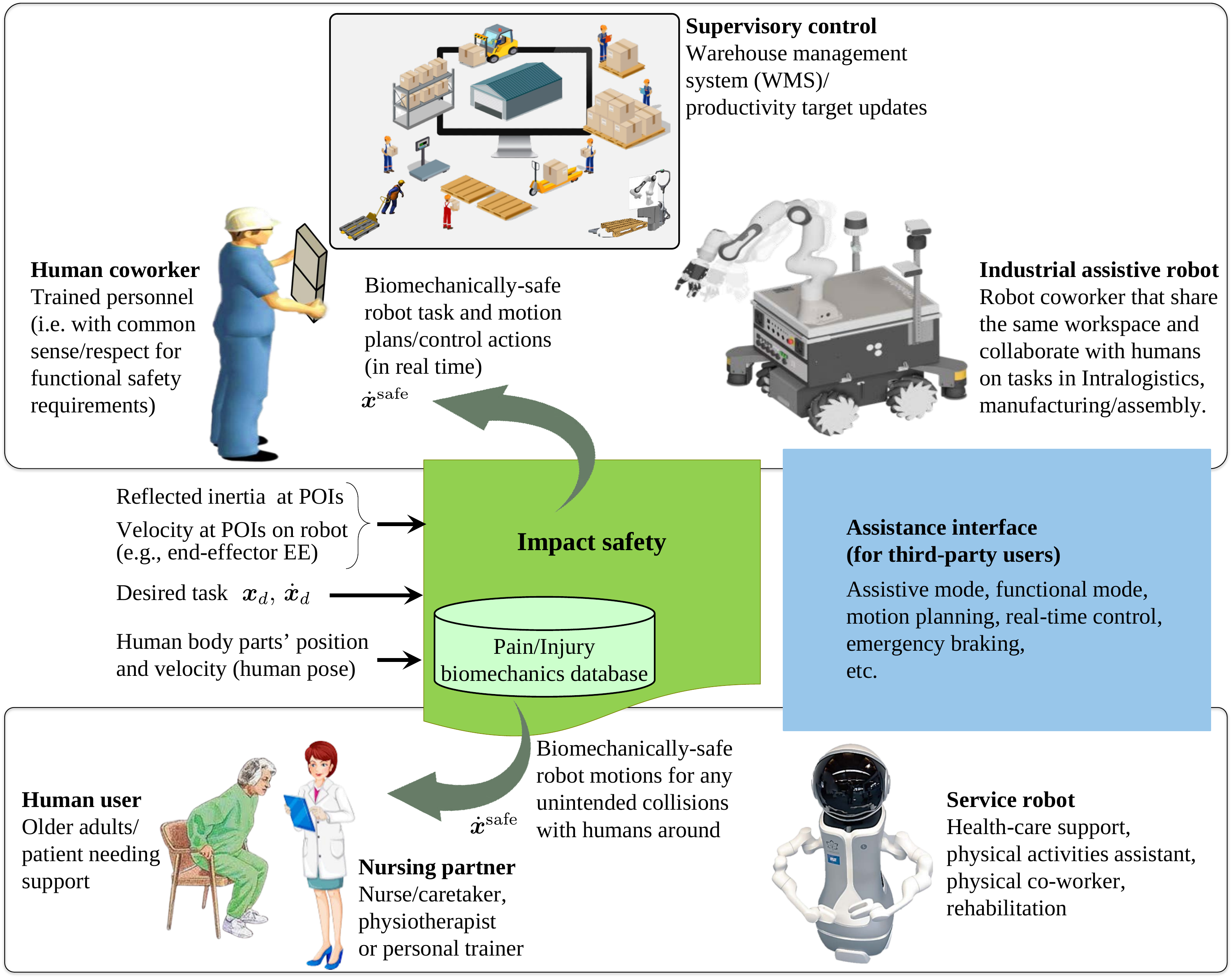}
		\caption[Impact safety layer]{Impact safety layer. Relying on human pain and injury information, this layer ensures that all physical robot-human contacts are biomechanically safe.}
		\label{fig:impact_safety_layer}
	\end{figure}

	A well-established injury analysis-based approach for addressing the safety requirements for stationary manipulator arms at the pre-collision phase was previously proposed in~\cite{haddadin2012understand}. For this injury biomechanics-based and impact data-driven approach, the so-called Safe Motion Unit (SMU) is the core tool for controlling the robot and some of the resulting dynamic collision parameters in a human-safe way. This systematic scheme was recently extended and generalized as a unified safety scheme for all floating-base robotic structures with branched manipulation extremities \cite{hamad2023modularize}. An abstraction of a generalized impact safety module is depicted in Fig.~\ref{fig:impact_safety_layer}.
	
	\subsubsection{Ergonomics}
	\label{sec:ergonomics}
	Typically, neurorehabilitation robots are programmed to interact autonomously with patients under clinician oversight (\ie, occupational and physical therapist) oversight such that safe and proper treatment is ensured \cite{hasson2023neurorehabilitation}. A major advantage of robot-assisted therapeutic treatment is the opportunity for accelerated patient recovery with frequency and duration of treatment being key factors \cite{colombo2000driven}. By precisely performing repetitive and mechanically power-consuming tasks, the robot drives the patient through ergonomically favorable positions during the whole training session. In contrast, any limitation of available degrees of freedom (DOF) during the robotic therapy can lead to changes in muscle activation patterns, negatively influencing its outcome \cite{schiele2006kinematic}. 
	
	Domestic or workplace ergonomics are addressed by performing risk assessments and analyzing human comfort during task execution. For this, ergonomists consider the worst posture achieved by taking measurements of the human's posture, either onsite or from video recordings. A comprehensive overview of the current state-of-the-art ergonomic human-robot collaboration in industrial settings was recently provided \cite{lorenzini2023ergonomic}. In their review, the authors not only investigated ergonomic assessment methodologies and available monitoring technologies for adapting robot control strategies online according to workers' distress and needs, but they also highlighted the most promising research themes and discussed state-of-the-art limitations and standing challenges. The main challenges lie in the cost-effectiveness of ergonomics monitoring, their comprehensive risk assessment methodologies, and the needed level of expertise to implement and maintain them. To handle the above issues, an ergonomically intelligent pHRI framework that includes smart and autonomous posture estimation, postural ergonomics assessment, and postural optimization was proposed in \cite{yazdani2021posture}. Furthermore, to overcome practical problems and risk assessment inaccuracies associated with commonly used discrete ergonomics models in performing postural optimization, differentiable and continuous variants of the famous and scientifically validated RULA and REBA\footnote{R(UL/EB)A: Rapid (Upper Limb/Entire Body) Assessment} ergonomics assessment models were learned via neural network regression \cite{yazdani2022dula}. As a result of a comparative study on the employed models and state-of-the-art developments for postural optimization in pHRI and teleoperation (cf.~Tab.~1 in \cite{yazdani2022dula}), DULA and DEBA\footnote{D(UL/EB)A: Differentiable (Upper Limb/Entire Body) Assessment} models were proposed as alternative differential models for improving both gradient-based and gradient-free posture optimizations.
	
\begin{figure}
		\centering
		\includegraphics[width=0.95\textwidth]{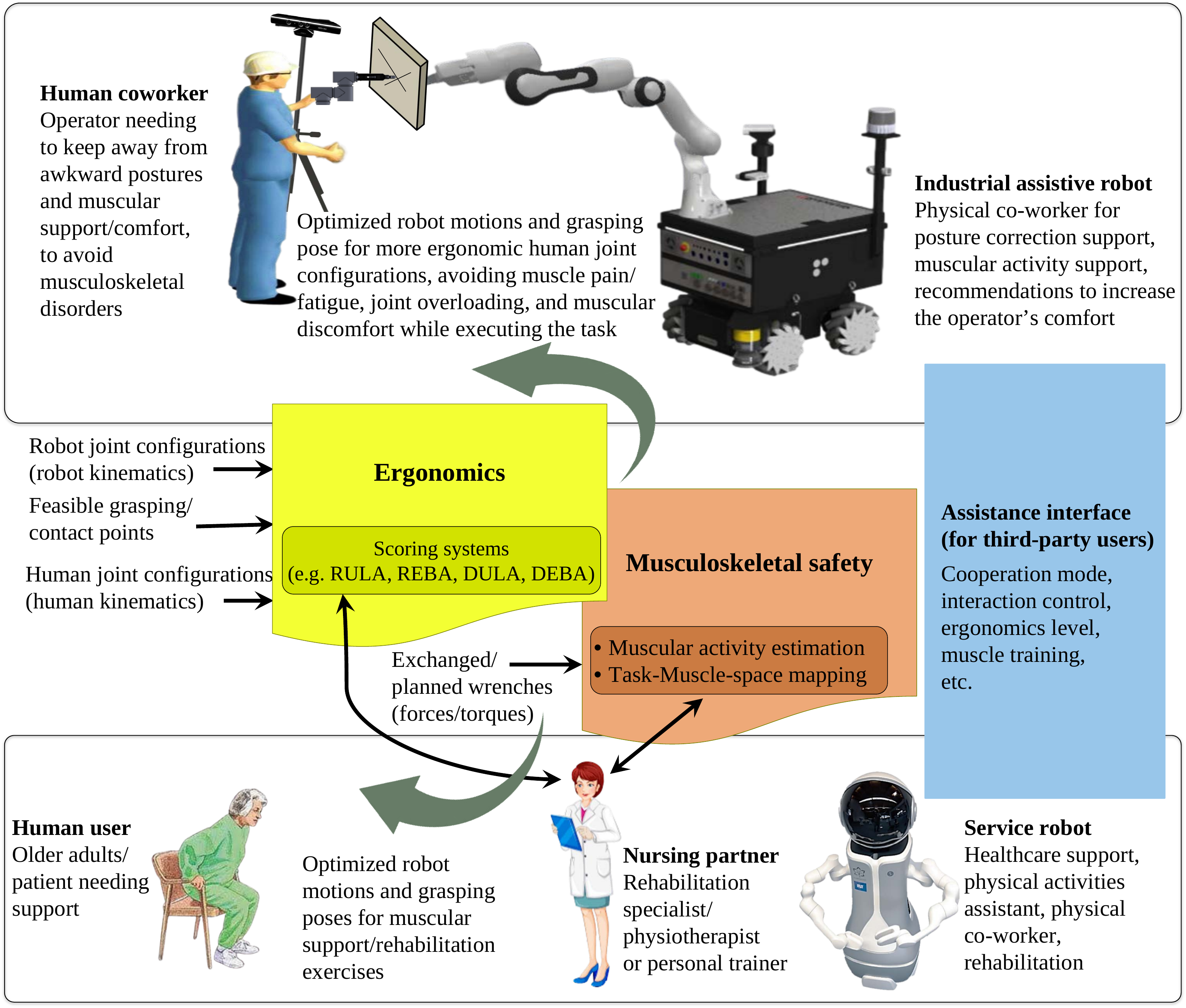}
		\caption[Ergonomics and musculoskeletal safety layers (combined)]{Ergonomics and musculoskeletal safety layers (combined). By optimizing the robot motion plans, the ergonomics layer ensures avoiding less ergonomic human postures during pHRI. On the other hand, by optimizing the robot motions and grasping poses, the musculoskeletal safety layer ensures avoiding the user's muscular discomfort/overloading during pHRI.} \label{fig:ergonomics_musculoskeletal_safety_layers_combined}
	\end{figure}

	By addressing static postural factors' influence, actions' repeatability, and experts' experience, ergonomic concepts are well-posed for high-level rapid task planning \cite{figueredo2020human}. A human-robot collaboration framework for improving ergonomic aspects of the human co-worker during power tool operations was proposed in \cite{kim2021human}. Nonetheless, ergonomic methods fail to address the impact and magnitude of larger forces and dynamic constraints in physical human-robot collaboration, which are better captured through muscular-informed metrics \cite{chen2018planning}. Building from existing literature, we propose a general abstraction for our ergonomics service module, as depicted in~Fig.~\ref{fig:ergonomics_musculoskeletal_safety_layers_combined}.   
	
	\subsubsection{Musculoskeletal safety}
	\label{sec:musculoskeletal_safety}
	In recent years, rehabilitation robotics has become indispensable for providing patients suffering from nervous system injuries with neurorehabilitation and movement therapy \cite{albert2012neurorehabilitation,iandolo2019perspectives}. These injuries include for example spinal cord injury, traumatic brain injury, or stroke. 
	For a recent, comprehensive systematic review on the effectiveness of robot-assisted rehabilitation techniques for patients recovering from musculoskeletal injuries or neurologic impairments, it is recommended that the reader consults  \cite{payedimarri2022effectiveness}.
	
	The application of robotic technologies in rehabilitation has progressed over the last few years. However, while the demand for medical rehabilitation services has been rapidly increasing \cite{kim2020factors}, the number of rehabilitation care providers continues to decrease annually %
	\cite{cieza2020global}. 
	Robotic medical devices are helpful for musculoskeletal therapy, where musculoskeletal symptoms such as myalgia, arthritis, postural instability, and fatigue are common disorders \cite{khan2020robotics}. These rehabilitation robots support regaining and improving the functional status, coordination, and independence of older adults \cite{sale2012use}. For instance, robot-aided locomotive treatment for stroke survivors and individuals coping with other neurologic impairments such as multiple sclerosis, Parkinson's disease, and spinal cord injury may involve either stationary, motion-based robots or exoskeletons \cite{chattoraj2023robotic}. Moreover, it was observed that impairments resulting from those diseases are becoming increasingly worrying for people under the age of 65 \cite{van2006susceptibility}. Besides walking-aid, typical daily-life activities where older adults or people with locomotive disorders need physical support during the \emph{Sit-to-Stand} and the \emph{Stand-to-Sit} transitional movements \cite{taylor2017assist}. \looseness=-1 
	
	Regarding industrial settings, a novel control approach was proposed in \cite{kim2017anticipatory} to alert and reduce a human partner's static joint torque overloading and consequent injury risks while executing shared tasks with a robot. An online optimization technique was employed for adjusting the robot trajectories to achieve more ergonomic human body poses, considering their stability, different workspaces (of robot and human), and task constraints. Furthermore, the problem of planning a robot configuration and shared object grasp during forceful human-robot collaboration is addressed in \cite{figueredo2021planning}. The proposed comfort planning framework aims to identify optimal robot configurations for jointly manipulating objects. This involves positioning the object in a way that minimizes the muscular effort exerted by the human and tailoring their collaborative actions accordingly. Additionally, the framework ensures the stability of the robot coworker during physical interaction. It enables the robot to shape human kinematics and musculoskeletal response while being agnostic to muscular activity estimation paradigms. 
	Building from existing literature, we propose a general abstraction for our musculoskeletal safety service module, as depicted in Fig.~\ref{fig:ergonomics_musculoskeletal_safety_layers_combined}. 
	
	\subsubsection{Perceived safety}
	\label{sec:perceived_safety}
	Although extensive research work has been carried out on physical safety in HRI scenarios, considerations of humans' expectations and affective state are often overlooked. In dynamic co-manipulation tasks, the robot may need to achieve higher velocities even when humans are present. To address the psychological safety of humans working in proximity to or directly with robots, an experimental setup was devised to examine the influence of robot velocity and robot-human distance on involuntary motion occurrence (IMO) caused by startle or surprise~\cite{kirschner2022expectable}. The relative frequency of IMO served as an indicator of potentially unsafe psychological situations for humans. The findings from these experiments were utilized to develop the Expectable Motion Unit (EMU) framework. The EMU ensures that IMO remains within a customizable probability range in typical HRI settings, thereby preserving psychological safety. This EMU is integrated into a comprehensive safety framework that combines psychological safety insights with the physical safety algorithm of the Safe Motion Unit (SMU). In a subsequent study, the efficiency of this psychologically-based safety approach in HRI was further enhanced by simultaneously optimizing both the Cartesian path and speed using  Model Predictive Control (MPC) such that the time taken to reach the target pose is minimized \cite{eckhoff2022mpc}. 
	
	To investigate the impact of robot motion and individual characteristics on users' perceived safety in HRI, a study was conducted involving human participants~\cite{nertinger2023influence}. The objective was to determine whether significant effects of human factors could be observed on IMO. The results of the study 
	revealed that direct human factors such as gender, age, profession, intention, technology anxiety, or curiosity to use did not significantly influence the occurrence of IMO. However, a noteworthy habituation effect was observed, indicating that participants became accustomed to the robot's motions quickly. 
	In the rather young subject sample which participated in the study of \cite{exeler2023influence}, only habituation showed a significant impact. Overall, those studies shed light on the interplay between robot motion, personal traits, and users' perceived safety in HRI, highlighting the importance of habituation and experimental design considerations. In \cite{tusseyeva2022perceived}, perceived safety in HRI for fixed-path and real-time motion planning algorithms was investigated based on arbitrary, physiological vital signs such as heart rate. The results emphasized that perceived safety is positively affected by habituation during the experiment and unaffected by previous experience.  
	A comprehensive discussion for increased perceived safety in HRI has been recently given in \cite{akalin2022you}, where the following guidelines are listed
	\begin{itemize}[topsep=2pt]
		\item Instead of seeking for the space of perceived safety, more focus should be put on objective metrics analysing a lack of perceived safety as significant indications for robot control schemes are mainly measurable under unsafe conditions;
		\item Regarding objective and subjective measures, robot-related and human-related factors should be treated together since the HRI process is bidirectional;
		\item The key influencing factors of perceived safety that should be considered in designing safe HRI are identified as comfort, experience/familiarity, predictability, sense of control, transparency, and trust;
		\item The consequences of robot-related factors, see for example \cite{akalin2019evaluating}, should not result in discomfort, lack of control, and user distrust, whereas the robot behaviours should be familiar, predictable, and transparent;
		\item Besides the interrelationship between the factors, individual human characteristics as well as emotional and physiological reactions should be considered for a better understanding of the source of safety perception. 
	\end{itemize}
	
	\subsubsection{Acceptance}
	\label{sec:acceptance}
	To improve industrial production tasks such as assembly, manufacturing, and intralogistics, human-robot collaboration (HRC) is instrumental. Even though there are apparent benefits of using robots in industrial workplaces, several barriers limit employing collaborative robots in the industry. These are not only related to strict safety regulations for physical human-robot collaboration (being the key show stopper for the investment from the employers' point of view), but also the workers' acceptance is crucial. In \cite{lotz2019you}, the main factors influencing the workers' acceptance of HRC are examined. In \cite{gombolay2015decision}, the authors hypothesized that giving human workers partial decision-making authority over a task allocation process for work scheduling maximizes both team efficiency and the desire of human team members to work with semi-autonomous robotic counterparts. Their experimental results indicated that workers prefer to be part of an efficient team rather than have a role in the scheduling process if maintaining such a role decreases their efficiency. 
	
	Acceptance is also a crucial factor for utilizing the potential of service robotics in facilitating domestic tasks, including required safety-critical measures. Moreover, meeting user expectations is essential for fostering trust between the human and the robot \cite{kellmeyer2018social,langer2019trust}. For instance, accepting an assistive robot to operate on-site in close physical interaction for medical examinations requires patient trust towards the robot. On the other hand, for human-in-the-loop (HIL) telemedicine, the presence of a human expert that remotely operates the robot can help the person trust the robot more and accept even its risky motions to perform the task. In \cite{nertinger2022acceptance}, a service robot was used to understand which outpatient-care tasks may be accepted by the subjects depending on their socio-demographics, beliefs, and level of robot autonomy. 
	
	\subsubsection{Personalization}
	\label{sec:personalization}
	Assistive robotics aims at providing users with continuous support and personalized assistance through appropriate interactions. Besides observing and understanding the changes in the environment to react promptly and behave appropriately, an intelligent assistive robot should be easy to handle, intuitive to use, ergonomic, and adaptive to human habits, individual usage profiles, and preferences. A personalized adaptive stiffness controller for pHRI tasks calibrated for the user's force profile was proposed for industrial applications in \cite{gopinathan2017user}. Its performance was validated in an extensive user study with multiple participants on two different tasks against a standard fixed controller. The results showed that the personalized approach was better regarding both performance gain and user preference, clearly pointing out the importance of considering both task-specific and human-specific parameters while designing control modes for pHRI. Furthermore, analyzing users' interaction force profiles, it was further confirmed that human and task parameters could be combined and quantified by considering the manipulability of a simplified human arm model. In \cite{mangin2022helpful}, a collaborative robotic system that is capable of assisting a human worker despite limited manipulation capabilities, incomplete task model, and partial environment observability was proposed. To achieve that, information from a high-level, hierarchical model is shared between the human and the robot, enabling transparent synchronization between the peers and mutual understanding of each other's plans.
	
	A socially assistive robotic system that can provide affordable personalized physical and cognitive assistance, motivation, and companionship with adaptable behaviour to its human user profile was first proposed in \cite{tapus2011user}. In subsequent work \cite{aly2014towards}, a fuzzy-based methodology was employed to investigate how matching the human and the robot personalities can influence their interaction. Furthermore, robot head-arm metaphoric gestures were generated automatically under different emotional states based on the prosodic cues of the interacting human. In \cite{umbrico2020holistic}, a novel cognitive approach that integrates ontology-based knowledge reasoning, automated planning, and execution technologies was recently proposed to endow assistive robots with intelligent features for performing personalized assistive tasks. These features include reasoning at different levels of abstraction, understanding specific health-related needs, and the ability to autonomously decide on how to act.
	
	\subsection{Additional middle-ware safety considerations}
	\label{sec:safety_consdierations}
	To adequately address the human diversity related to both safety and security, some customization and individualization are necessary. In terms of physical safety, for instance, investigations on scaling issues (age and gender effects on material properties) and statistical methods have been conducted, see, e.g., \cite{kent2005structural,lillie2016evaluation,seeman2001during}, 
	for estimating the human injury risk curves, using various anthropomorphic test devices (ATDs) and mathematical models of the human body \cite{crandall2011human}. International anthropometric data for the workplace and machinery design can be found in, \eg~\cite{jurgens1998international}, and the corresponding technical report \cite{fubini2010international}. 
	On the other hand, employing physiological measurements to perform online assessment of operators' mental states is crucial in HRI. To progress towards interactive robotic systems that would dynamically adapt to operators' affective states, in \cite{drougard2018physiological}
	operator's recorded physiological data streams were analyzed to assess the engagement during HRI and the impact of the robot's operative mode (autonomous versus manual). Furthermore, a software framework that is compatible with both laboratory and consumer-grade sensors, while it includes essential tools and processing algorithms for affective state estimation, was recently proposed in \cite{kothig2020hri} to support real-time integration of physiological adaptation in HRI.
	
	\section{Safety-as-a-Service: Implementation Prospects}
	\label{sec:implementation_prospects}
\begin{figure}
		\centering
		\includegraphics[width=0.95\textwidth]{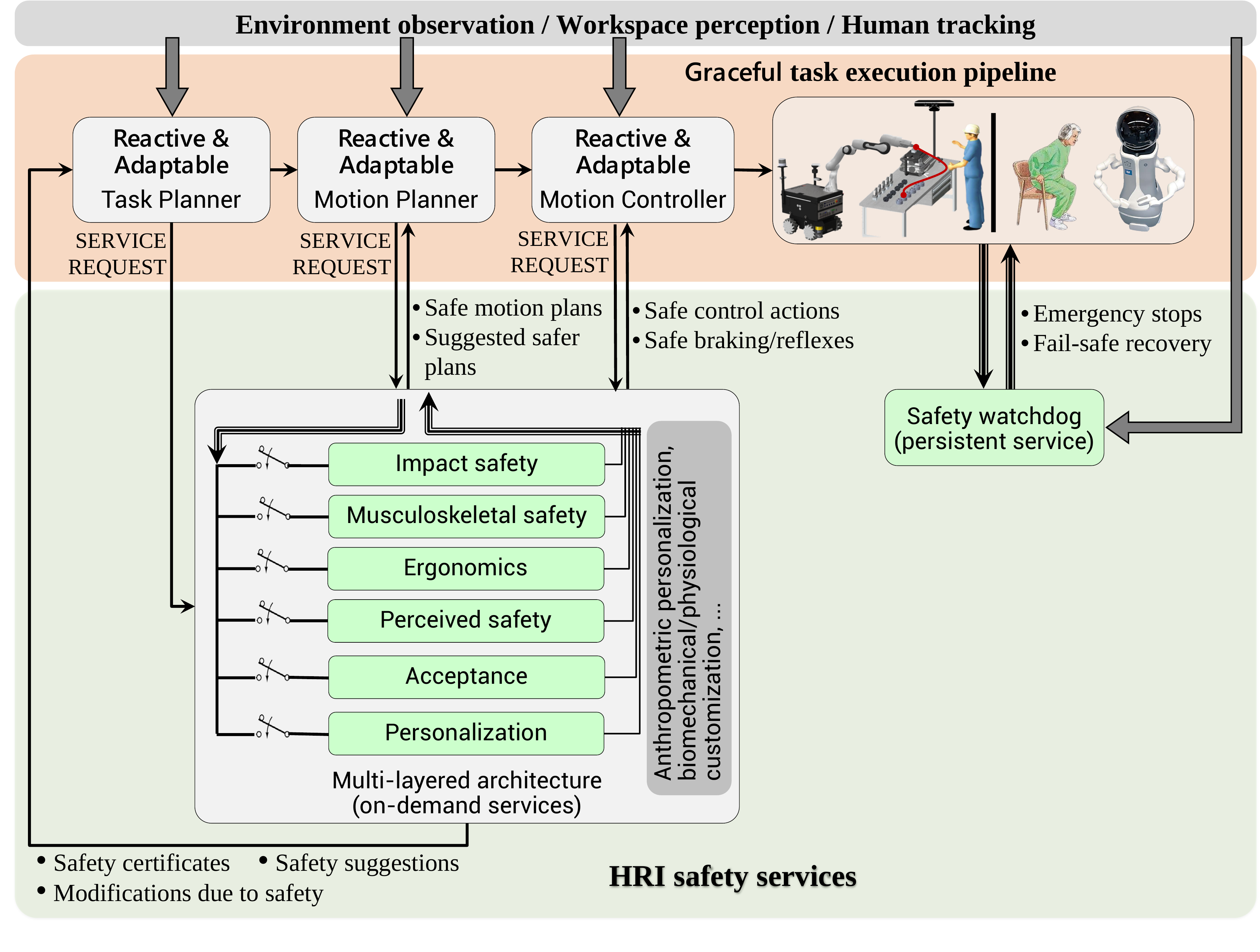}
		\caption[Safety-as-a-service concept]{Safety-as-a-service concept for GS-HRI.}
		\label{fig:safety_as_a_service_concept}
	\end{figure}

	The schematics shown in Fig.~\ref{fig:safety_as_a_service_concept} demonstrate our proposed concept of providing different safety services upon request at different stages of the graceful task execution pipeline. The latter is obtained by redesigning motion controllers, motion planners, and task planners of the user-defined collaborative robotic task execution pipeline. The aim is to satisfy the reactivity and adaptivity requirements imposed by the safety layers for a GS behaviour. Furthermore, the functionality of each safety layer is encoded as an on-demand service. In contrast, critical safety aspects are ensured via persistent (\ie, always on) services such as emergency braking or safe fault recovery operation modes. 
	
\begin{figure*}
		\centering
		\includegraphics[width=0.99\textwidth]{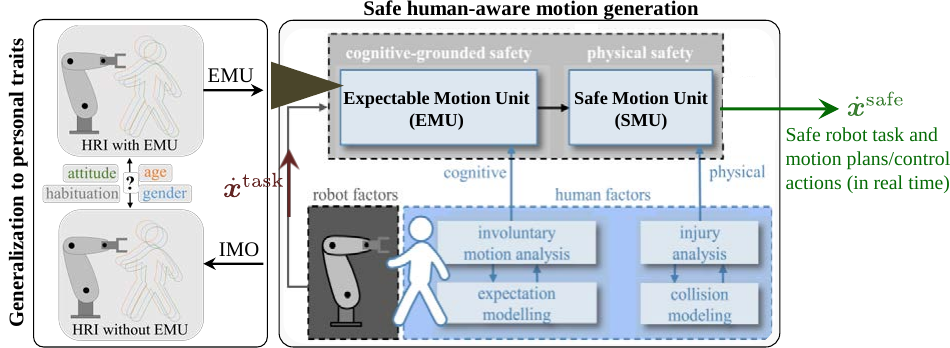}
		\caption[Including human factors in EMU-SMU safe HRI framework \cite{kirschner2022expectable,nertinger2023influence}.]{Including human factors in the EMU-SMU safe HRI framework \cite{kirschner2022expectable,nertinger2023influence}.}
		\label{fig:personal_traits_EMU_SMU_merge}
	\end{figure*}

	To implement various safety services, the so-called generalized Safe Motion Unit (gSMU) framework can be adopted as the underlying safety-certifiable scheme for providing biomechanically-safe robot motions \cite{hamad2023modularize}, with the possibility to include additional robot payload \cite{hamad2019payload} and predictable braking strategies \cite{hamad2023fast}. 
	Simultaneous consideration of human factors, especially experience with robots/habituation, which potentially influences the humans' perceived safety for varying robot factors \cite{nertinger2023influence}, can be achieved by including them in the EMU-SMU framework \cite{haddadin2012understand,kirschner2022expectable}, see Fig.~\ref{fig:personal_traits_EMU_SMU_merge}.  

\section{Conclusion and Future Work} 
\label{sec:conclusion_future_work}
This work presented an integrated multi-layered architecture to simultaneously tackle safety issues as well as gracefulness requirements of human-friendly robots in HRI scenarios. Based on a focused literature review, we identified various physical and cognitive HRI safety layers and emphasized notable studies discussing each and their corresponding findings. Furthermore, we suggested the \emph{safety-as-a-service} concept for formalizing how to address the requirements of each HRI safety aspect concurrently while adapting the collaborative task execution pipeline for graceful robot task execution. Then, we discussed an example that shows some promising integration work along the suggested direction.  

For future research and developmental work, we will detail some crucial architectural aspects such as prioritization of safety features that generate service requests, smooth switching and management of multiple ones concurrently, hierarchical rules needed to handle conflicts that may arise at the output level from different layers, as well as elaboration of the middleware considerations. We also plan to study the possibility of extending the safety assessment methodology proposed in \cite{vicentini2019safety} to cover the cognitive aspects, such that it can be employed for formal verification of the proposed multi-layered HRI safety architecture. 
Moreover, a comprehensive user study in industrial and service robot settings with a heterogeneous subject sample (including a broad range of persons with different experiences, ages, and genders) is required. 
Further, as users are able to adjust their expectations of the robot's behaviour quickly (habituation), possible efficiency enhancements of the human-robot teams are feasible. Also, the effect of unfulfilled expectations on following interactions needs to be analyzed by means of subjective and objective measures. \looseness=-1

\normalem
\renewcommand{\baselinestretch}{0.8}

\bibliographystyle{splncs03_unsrt}
\bibliography{safety_as_a_service}

\end{document}